\documentclass[sigconf]{acmart}

\usepackage{multirow}
\usepackage{subfigure}
\usepackage{bm}

\acmSubmissionID{156}

\begin{document}

\title{Joint Learning of Local and Global Features for Aspect-based Sentiment Classification}


\author{Hao Niu}
\affiliation{%
  \institution{Shanghai Key Laboratory of Data Science, School of Computer Science, Fudan University}
  \city{Shanghai}
  \country{China}}
\email{hniu18@fudan.edu.cn}

\author{Yun Xiong}
\affiliation{%
  \institution{Shanghai Key Laboratory of Data Science, School of Computer Science, Fudan University}
  \city{Shanghai}
  \country{China}}
\email{yunx@fudan.edu.cn}
\authornote{Corresponding author}

\author{Xiaosu Wang}
\affiliation{%
  \institution{Shanghai Key Laboratory of Data Science, School of Computer Science, Fudan University}
  \city{Shanghai}
  \country{China}}
\email{xswang19@fudan.edu.cn}

\author{Philip S. Yu}
\affiliation{%
  \institution{Department of Computer Science,
University of Illinois at Chicago}
  \country{USA}}
\email{psyu@cs.uic.edu}




\renewcommand{\shortauthors}{Trovato and Tobin, et al.}

\begin{abstract}
  Aspect-based sentiment classification (ASC) aims to judge the sentiment polarity conveyed by the given aspect term in a sentence. The sentiment polarity is not only determined by the local context but also related to the words far away from the given aspect term. Most recent efforts related to the attention-based models can not sufficiently distinguish which words they should pay more attention to in some cases. Meanwhile, graph-based models are coming into ASC to encode syntactic dependency tree information. But these models do not fully leverage syntactic dependency trees as they neglect to incorporate dependency relation tag information into representation learning effectively. In this paper, we address these problems by effectively modeling the local and global features. Firstly, we design a local encoder containing: a Gaussian mask layer and a covariance self-attention layer. The Gaussian mask layer tends to adjust the receptive field around aspect terms adaptively to deemphasize the effects of unrelated words and pay more attention to local information. The covariance self-attention layer can distinguish the attention weights of different words more obviously. Furthermore, we propose a dual-level graph attention network as a global encoder by fully employing dependency tag information to capture long-distance information effectively. Our model achieves state-of-the-art performance on both SemEval 2014 and Twitter datasets.
\end{abstract}

\maketitle




\section{Introduction}

Aspect-based sentiment classification (ASC) is a fine-grained sentiment classification task in the field of sentiment analysis. The goal is to predict the sentiment polarity (e.g., \textsc{positive}, \textsc{negative}, \textsc{neutral}) for one or more aspect terms discussed in a given sentence. For example, the sentence: \emph{great food but the service was dreadful} \cite{huang-carley-2019-syntax}, would be assigned with \textsc{positive} polarity for the aspect term \emph{food} and with a \textsc{negative} polarity for the aspect term \emph{service}. Since the two aspect terms express opposite sentiment polarities, it is inappropriate to allocate an overall sentiment polarity for a sentence. In this respect, ASC provides more elaborate analyses than sentence-level sentiment analysis.

In recent years, contextualized language models are used in the ASC task. For example, AEN-BERT \cite{DBLP:journals/corr/abs-1902-09314} uses BERT to encode both context \& aspect sequences and employs an attention mechanism to explore semantic information between aspect terms and context words. On the other hand, LCF-BERT \cite{DBLP:conf/acl/PhanO20} holds the belief that valuable information is more likely to appear around aspect terms and intends to localize context information around the given aspect term by down-weighting the contributions of words that are far away from the given aspect term. Though this approach is beneficial to get rid of the impact of irrelevant words far away from the given aspect term, it also leads to missing some crucial information. Moreover, LCF-BERT adjusts the local context receptive field around the given aspect term by utilizing a context masking matrix generated through setting hyper-parameters. However, this approach is insufficient and inflexible to encode the local semantic information around the given aspect term. Meanwhile, LCF-BERT does not sufficiently model global information since it ignores syntactic dependency relations that long-distance semantic information conveys.

The self-attention mechanism proposed by \cite{DBLP:journals/corr/VaswaniSPUJGKP17} is used in ASC \cite{DBLP:conf/ijcai/MaLZW17, DBLP:conf/emnlp/ChenSBY17, DBLP:conf/emnlp/FanFZ18}. However, the original self-attention mechanism cannot perform well enough for ASC. In some cases, the original self-attention is hard to identify valuable words, leading to a wrong sentiment polarity prediction. 

Besides, intuitively, to assign proper sentiment polarities to aspects, directly connecting aspect terms with their relevant opinion words would be more effective. Therefore, graph-based methods are proposed to tackle this task. Graph neural networks (GNNs) are explored to encode syntactic dependency trees \cite{zhang-etal-2019-aspect, sun-etal-2019-aspect, huang-carley-2019-syntax} to obtain syntactic dependency representations. However, these methods do not utilize the syntactic dependency relation tags that convey much information, such as \emph{nsubj} (nominal subject) and \emph{dobj} (direct object). Besides, though BERT-RGAT\cite{DBLP:conf/acl/WangSYQW20} considers dependency relation tags firstly, this model does not excavate edge information sufficiently by just considering edge types whereas neglecting the information of what the edges convey.

In summary, the existing efforts face the following challenges: (1) failure to make full use of local and global information effectively; (2) the original self-attention mechanism does not perform well enough; (3) the tag information of the syntactic dependency tree is not used effectively.

To handle these challenges we mentioned above, we propose a novel model named \emph{Gaussian mask Distinguishable and Dual-level attention structure (GDD)} to encode both local and global semantic information concurrently. GDD consists of four major components: input representation layer, local encoder, global encoder, and output layer. Firstly, we obtain the representations of input sentences and their corresponding dependency relation tags from the input representation layer. Then, we feed these representations into local and global encoders to capture local and global features, respectively. Unlike LCF-BERT, we propose a novel local encoder equipped with a \emph{Gaussian mask layer} and a \emph{covariance self-attention layer} to adjust local context receptive fields adaptively and make attention weights more distinguishable to encode local information effectively. Moreover, as a supplement to local information and preventing missing some long-distance information, we also plan to excavate global information. We construct an \emph{Aspect-word Interactive Graph (AWIG)} to directly connect the given aspect term and the words, whether they are one-hop or multi-hop connections with the aspect term in the original dependency parse tree generated by the Biaffine Parser \cite{DBLP:conf/iclr/DozatM17}. In this way, we put the long-distance words around aspect terms and directly establish the relations between aspect terms and potential long-distance opinion words, which is good for capturing global information. In addition, our AWIG is edge attributed, and we assign the corresponding dependency relation tag to each edge. To encode the constructed graph with attributed edges, we propose a novel \emph{dual-level graph attention network (DGAT)} to capture the global information that AWIG contained. Lastly, we concatenate the outputs from local and global encoders as the final representation of each input case and employ a fully connected layer to make predictions. The contributions of our work can be highlighted as:
\begin{itemize}
    \item \textbf{Gaussian mask:} we propose a \emph{Gaussian mask layer} assembled in the local encoder to adjust the local context receptive fields around the aspect terms adaptively.
    \item \textbf{Covariance Self-attention:} we propose a \emph{covariance self-attention layer} constructed in the local encoder to make attention scores more distinguishable.
    \item \textbf{Dual-level Attention:} we design a novel graph neural network \emph{dual-level graph attention network (DGAT)} by applying dependency relation tag information comprehensive to excavate valuable long-distance information.
\end{itemize}

\section{Related Work}
\subsection{Attention-based Models}
Most recent researchs on aspect sentiment classification utilize attention-based models to explore relative context words around aspect terms. Since opinion words usually appear near to aspect terms, focusing on more local context information surrounding aspect terms would be beneficial in predicting correct sentiment polarity. For example, ATAE-LSTM\cite{DBLP:conf/emnlp/WangHZZ16} introducing attention-based LSTM is the first attempt to apply attention mechanism to ASC. MGAN \cite{DBLP:conf/emnlp/FanFZ18} utilizes a multi-grained attention network, which contains fine-grained and coarse-grained attention to capture information surrounding aspect terms.
RPAEN-BERT \cite{DBLP:journals/kais/WuXGLYW21} fine-tunes BERT and employs a relative position encode layer to integrate the relative position information of aspect terms to make predictions. LCF-BERT\cite{DBLP:conf/acl/PhanO20} employs a context masked matrix to localize sentiment context information by setting a hyper-parameter syntactic relative distance (SRD). Although the approach of LCF-BERT can eliminate the influence of unrelated words, this method is too inflexible by setting hyper-parameters manually to adjust the width of the local context receptive field.

\begin{figure*}
\centering
\includegraphics[width=0.83\textwidth]
{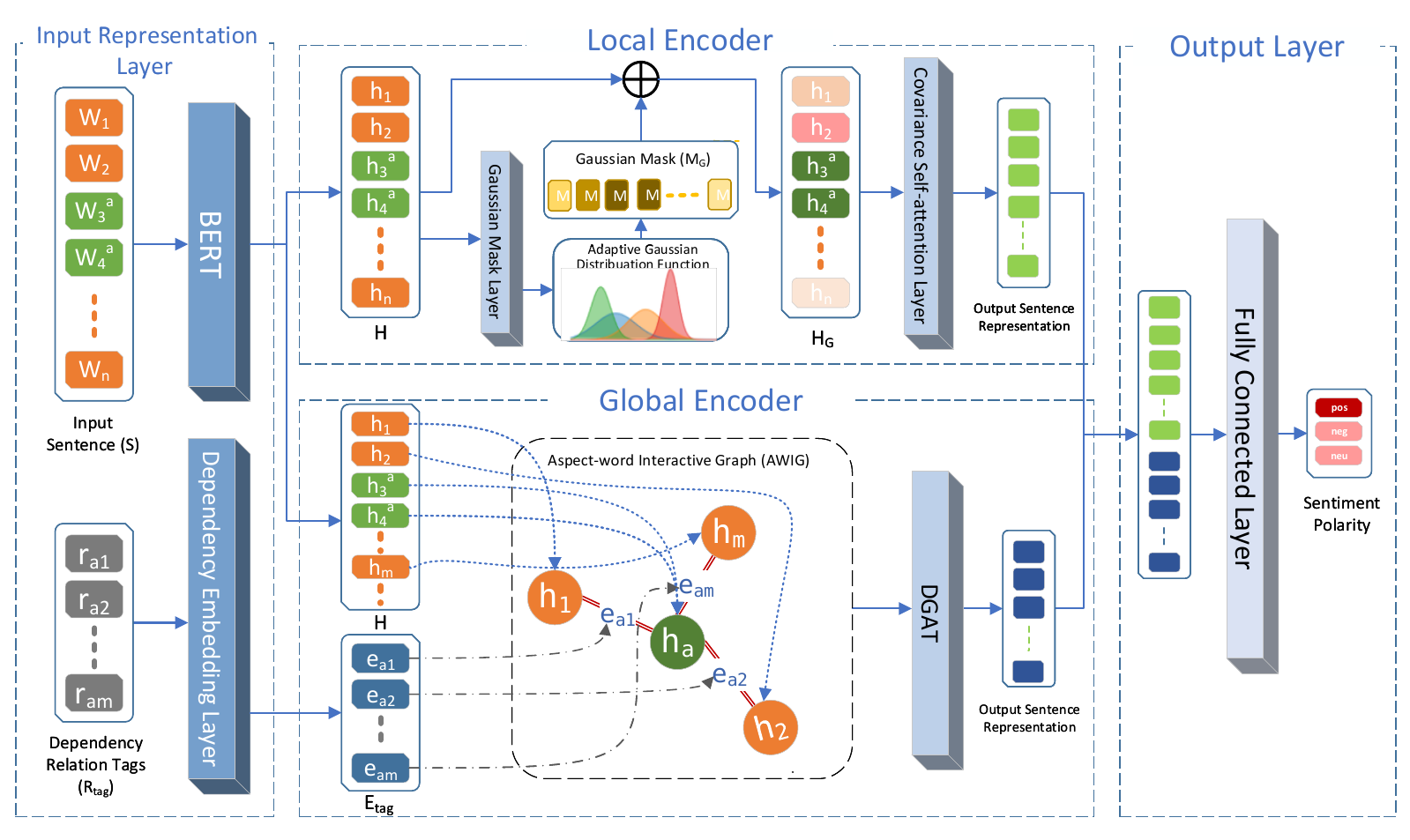}
\caption{Overall architecture of proposed GDD.}
\label{fig:GDD}
\end{figure*}
\subsection{Graph-based Models}
Recently, the methods incorporating graph neural networks (GNNs) with syntactic dependency trees have been introduced in the aspect-based sentiment classification field. For example, ASGCN\cite{zhang-etal-2019-aspect} and CDT\cite{sun-etal-2019-aspect} introduce graph convolutional networks (GCNs) into aspect sentiment classification to encode syntactic dependency tree-based graph structure and merge it with other features to make predictions. For a similar purpose, Huang\cite{huang-carley-2019-syntax}  and Wang\cite{DBLP:conf/acl/WangSYQW20} propose to use the graph attention network (GAT), as well as its derivatives, to encode the syntactic dependency relationship between non-adjacent words. Although BERT-RGAT \cite{DBLP:conf/acl/WangSYQW20} is the first to take dependency relation tags into consideration, this model could not excavate edge information sufficiently by just considering edge types whereas neglecting the information of what the edges convey. Therefore, all these approaches mentioned above cannot sufficiently utilize dependency relation tags, which are valuable when judging the proper sentiment polarity.


\section{Proposed Method}
\label{sec:length}
Aspect-based sentiment classification can be formulated as follows: given a sentence $S$ with $n$ words $\{w_1, w_2, ..., w_n\}$ and the given aspect term $A$ = $\{w_i^a, w_{i+1}^a, ..., w_{i+\tau}^a\}$ with $\tau \in [1, n)$ is a substring of $S$. The superscript $a$ represents the given aspect term. The task aims at predicting the sentiment polarity $y \in $ $\textsc{\{positive, neutral, negative\}}$ expressed on the given aspect term in the given sentence. 

In order to solve the task mentioned above, we propose a novel model called GDD to make full use of local and global information concurrently and enhance the performance of ASC. In this section, we describe our proposed model in detail. The overall architecture of our model is shown in Figure \ref{fig:GDD}. There are four major components: (1) input representation layer, which encodes words and dependency relation tags into vector representations; (2) local encoder, which encodes local context information around the given aspect term adaptively; (3) global encoder, which captures effective global information by utilizing syntactic dependency trees; (4) output layer, which uses the final representation to calculate the probabilities for different sentiment polarities.

\subsection{Input Representation Layer}
For a given input sentence $S = \{w_1, w_2, ..., w_n\}$ with the aspect term $A = \{w_i^a, w_{i+1}^a, ..., w_{i+\tau}^a\}$ included, we first apply the pre-trained BERT \cite{DBLP:conf/naacl/DevlinCLT19} as the contextual encoder to extract hidden contextual representation $\bm{H}$, and this representation contains two parts: word embeddings $\{\bm{h_1}, \bm{h_2}, ..., \bm{h_n}\}$ and the given aspect embedding $\{\bm{h_i^a}, \bm{h_{i+1}^a}, ..., \bm{h_{i+\tau}^a}\}$. The hidden contextual representation output from pre-trained BERT is divided into two streams and fed into the local and global encoder, respectively. Additionally, to utilize dependency relation tag information, we employ a trainable embedding lookup table as a dependency embedding layer. The dependency relation tags $R_{tag} = \{r_{a1}, r_{a2}, ..., r_{am}\}$ are generated by the Biaffine Parser \cite{DBLP:conf/iclr/DozatM17} and modified following the setting of the second paragraph of the section \ref{section: section 3.3.1}. Then, we feed them into the dependency embedding layer to initialize dependency relation representations $\bm{E_{tag}} = \{\bm{e_{a1}}, \bm{e_{a2}}, ..., \bm{e_{am}}\}$, where $m$ is the number of the dependency relation tags for each input case. Then, the dependency relation representations are fed into the global encoder.

\subsection{Local Encoder}
To encode local context information effectively, we propose a novel local encoder containing two layers: a Gaussian mask layer and a covariance self-attention layer. The Gaussian mask layer can adjust the local context receptive field around the given aspect term, and the covariance self-attention layer is helpful to capture valuable words effectively. We detail the local encoder in this section.
\subsubsection{Gaussian Mask Layer}
To adjust local context receptive field in the surrounding area of aspect terms adaptively, we propose a adaptive module to learn a set of Gaussian mask to regulate the width of each local context. Formally, we design a multi-layer neural network to learn the standard deviation $\sigma$ to control the shape of Gaussian Distribution, $\sigma$ is computed as follow:
\begin{equation}
    \sigma = \delta(\bm{W_2}(relu(\bm{W_1 H} + \bm{b_1})) + \bm{b_2}),
\end{equation}
where $\delta$ is softplus function, $\bm{H}$ is the hidden contextual representation of the input sentence $S$ from BERT, and $\bm{W_1}, \bm{W_2}, \bm{b_1}$, and $\bm{b_2}$ are learnable weights and biases, respectively.  For the sake of simplicity, the mean $\mu$ of each Gaussian Distribution is predefined as zero. Gaussian Distribution is defined as follow:
\begin{equation}
    GK(x) = \frac{1}{\sigma\sqrt{2\pi}} e ^{-\frac{1}{2} \cdot (\frac{x}{\sigma})^2}.
\end{equation}
After acquiring the specific Gaussian Distribution $GK$, we generate the Gaussian mask vector $\bm{M_G}$ by setting the values of $[B_{i}^a, B_{i+1}^a, ..., B_{i+\tau}^a]$ to $GK(0)$, where the superscript $a$ represents the given aspect term, and sampling at equal intervals on both sides of zero on this specific Gaussian Distribution $GK$ to get the values of $[\dotsb B_{i-1}]$ and $[B_{i+\tau+1} \dotsb]$. The sample interval is a hyper-parameter, and the length of $\bm{M_G}$ is equal to the length of the input sequence, 
\begin{equation}
    \bm{M_{G}} = [\dotsb B_{i-1}, B_{i}^a, B_{i+1}^a, \dotsb, B_{i+\tau}^a, B_{i+\tau+1} \dotsb].
\end{equation}
After getting the Gaussian mask vector $\bm{M_G}$, the local context feature $\bm{H_G}$ is computed as follow.
\begin{equation}
    \bm{H_G} = \bm{M_G} \odot \bm{H},
\end{equation}
where $\odot$ denotes the Hadamard product to mask out the hidden contextual representation $\bm{H}$ by applying the learned Gaussian mask vector $\bm{M_G}$.

\subsubsection{Covariance Self-attention}
To enhance the performance of the self-attention mechanism, we would like to make the attention weights of different context words as distinguishable as possible. Inspired by \cite{DBLP:conf/eccv/YinYCLZLH20}, we propose a covariance self-attention mechanism to encode local context feature $\bm{H_G}$ of the given aspect term. The original self-attention \cite{DBLP:journals/corr/VaswaniSPUJGKP17} computes the attention score matrix between features of each two words. It computes as follow:
\begin{equation}
    Att(\bm{Q},\bm{K},\bm{V}) = softmax(\frac{\bm{Q}^\top\bm{K}}{\sqrt d_k})\bm{V},
\end{equation}
where $\bm{Q}$, $\bm{K}$, and $\bm{V}$ are queries, keys, and values matrices, and $d_k$ is the dimension of keys and queries. To amplify the difference between any word pairs and make their attention weights more obvious, we modify the original self-attention mechanism by subtracting a mean term on both $\bm{Q}$ and $\bm{K}$. Specifically, the covariance self-attention mechanism computes attention scores between any word pairs: $(\bm{Q}-\bm{U_Q})^\top(\bm{K}-\bm{U_K})$, where $\bm{U_Q} \!=\! \frac{1}{N}\sum_{j=1}^N{\bm{q_j}}$, $\bm{U_K}\!=\!\frac{1}{N}\sum_{\xi=1}^N{\bm{k_\xi}}$ are averaged query and key embedding over all words respectively, and $N$ is the number of words. The covariance self-attention computes as follow:
\begin{equation}
    Att(\bm{Q},\bm{K},\bm{V})=
softmax\left(\frac{(\bm{Q}-\bm{U_Q})^\top(\bm{K}-\bm{U_K})}{\sqrt{d_k}}\right)\bm{V}.
\end{equation}
We demonstrate theoretical analyses hidden behind the covariance self-attention mechanism. In the following proposition, we set an optimization objective \cite{DBLP:conf/eccv/YinYCLZLH20} to make attention weights differentiable as much as possible. Intuitively, maximizing the optimization objective denotes maximizing the difference between two word pairs' similarities, which matches our original purpose.

\paragraph{\textbf{Proposition:}} The optimal solution of the following optimization objective \emph{O} is $\theta^*\!=\!\frac{1}{N}\sum_{j=1}^N{\bm{q_j}}$ and $\varphi^*\!=\!\frac{1}{N}\sum_{\xi=1}^N{\bm{k_\xi}}$.
\begin{multline}
    arg\ max_{\theta, \varphi} 
    \frac{\sum_{j,\xi,\eta\in N} ((\bm{q_j}\! -\! \theta)^\top(\bm{k_\xi}\! -\! \varphi) \!- \!(\bm{q_j}\! -\! \theta)^\top(\bm{k_\eta}\! -\! \varphi))^2}{\sum_{j \in N ((\bm{q_j} - \theta)^\top(\bm{q_j} - \theta)) \cdot \sum_{\xi,\eta \in N} ((\bm{k_\xi} - \bm{k_\eta})^\top(\bm{k_\xi} - \bm{k_\eta}))}} \\
    \!+ \frac{\sum_{\xi,j,p\in N} ((\bm{k_\xi} \!-\! \varphi)^\top(\bm{q_j}\! -\! \theta) \!-\! (\bm{k_\xi} \! - \! \varphi)^\top(\bm{q_p} \! -\! \theta))^2}{\sum_{\xi \in N ((\bm{k_\xi} - \varphi)^\top(\bm{k_\xi} - \varphi)) \cdot \sum_{j,p \in N} ((\bm{q_j} - \bm{q_p})^\top(\bm{q_j} - \bm{q_p}))}}.
\end{multline}

\paragraph{\textbf{Proof Sketch :}} In order to optimize above mentioned objective, we utilize 
the nature of Rayleigh Quotient and Lagrangian Multiplier Method to calculate the optimal $\theta^*$ and $\varphi^*$, we can prove $\theta^*\!=\!\frac{1}{N}\sum_{j=1}^N{\bm{q_j}}$, $\varphi^*\!=\!\frac{1}{N}\sum_{\xi=1}^N{\bm{k_\xi}}$ is the optimal solution for this objective.

\paragraph{\textbf{Detailed Proof of Proposition:}} The optimization objective function $O(\theta, \varphi)$ in Eq.(8) can be reexpressed as follow:
\begin{align}
    O(\theta, \varphi) = & \frac{\sum_{j \in N}(\bm{q_j} - \theta)^\top \bm{\Phi} (\bm{q_j} - \theta)}{\sum_{j \in N}(\bm{q_j} - \theta)^\top (\bm{q_j} - \theta)} \notag \\ & + \frac{\sum_{\xi \in N}(\bm{k_\xi} - \varphi)^\top \bm{\Omega} (\bm{k_\xi} - \varphi)}{\sum_{\xi \in N}(\bm{k_\xi} - \varphi)^\top (\bm{k_\xi} - \varphi)},
\end{align}
where
\begin{equation}
\bm{\Phi} = \frac{\sum_{\xi,\eta \in N} (\bm{k_\xi} - \bm{k_\eta})(\bm{k_\xi} - \bm{k_\eta})^\top}{\sum_{\xi,\eta \in N}(\bm{k_\xi} - \bm{k_\eta})^\top(\bm{k_\xi} - \bm{k_\eta})}, 
\end{equation}
\begin{equation}
\bm{\Omega} = \frac{\sum_{j,p \in N} (\bm{q_j} - \bm{q_p})(\bm{q_j} - \bm{q_p})^\top}{\sum_{j,p \in N}(\bm{q_j} - \bm{q_p})^\top(\bm{q_j} - \bm{q_p})}.
\end{equation}
$O(\theta, \varphi)$ is composed by two Rayleigh Quotient form items. We utilize Lagrangian Multiplier Method to calculate the optimal solution of this objective function. Firstly, we prove that $\theta^*\!=\!\frac{1}{N}\sum_{j=1}^N{\bm{q_j}}$ is the optimal solution of the first item for the above optimization objective. The first item of such objective could be expressed as follow:
\begin{equation}
R(\theta) = \frac{\sum_{j \in N}(\bm{q_j} - \theta)^\top \bm{\Phi} (\bm{q_j} - \theta)}{\sum_{j \in N}(\bm{q_j} - \theta)^\top (\bm{q_j} - \theta)}.
\end{equation}
According to the Rayleigh Quotient and Lagrangian Multiplier Method nature, we would like to calculate the extreme point for Eq.(12). Obviously, proportional scaling of $(\bm{q_j} -  \theta)$ does not change the value of $R(\theta)$. Thus we restrict the value of $(\bm{q_j} - \theta)^\top (\bm{q_j} - \theta)$ to be equal to 1 and define a Lagrangian function as follow:
\begin{align}
\widetilde{L}(\theta, \lambda) = & \sum_{j \in N}(\bm{q_j} - \theta)^\top \bm{\Phi} (\bm{q_j} - \theta) \notag \\ 
&- \lambda((\bm{q_j} - \theta)^\top (\bm{q_j} - \theta) - 1).
\end{align}
By solving this equation, we have:
\begin{equation}
\sum_{j=1}^N(\bm{q_j} - \theta^*) = 0.
\end{equation}
The optimal $\theta^*$ is
\begin{equation}
\theta^* = \frac{1}{N}\sum_{j=1}^N \bm{q_j}.
\end{equation}
Similarly, the optimal $\varphi^*$ can be solved as
\begin{equation}
\varphi^* = \frac{1}{N}\sum_{j=1}^N \bm{k_j}.
\end{equation}

\subsection{Global Encoder}
In this section, to capture global context information, we first feed the hidden contextual representations and the dependency relation representations into the global encoder. Secondly, we build an edge attributed graph called aspect-word interactive graph (AWIG) to represent the relations between the given aspect term and corresponding words. Then, we propose a dual-level graph attention network (DGAT) to utilize the information of the constructed graph more effectively. Compared with other graph-based models for ASC, the novelty of our proposed model is to pay more attention to edge features, which is left in the basket by other works. And finally, we feed the constructed graph attached with its node and edge features into DGAT to obtain the final representation of the given aspect term for each input case. Below we describe the global encoder in detail.

\subsubsection{Aspect-word Interactive Graph (AWIG)}
\begin{figure}
\centering
\includegraphics[width=0.45\textwidth]{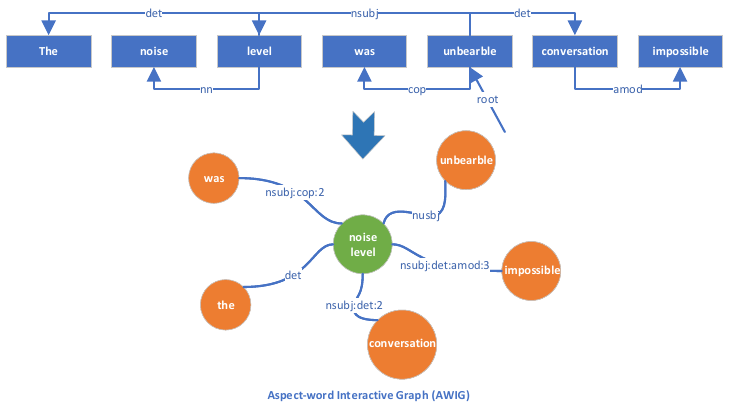}
\caption{Construction of an aspect-word interactive graph (AWIG) (bottom) from an original dependency parse tree (top). The aspect term is \textit{noise level} with the sentiment polarity \textsc{negative}. }
\label{fig:comp}
\end{figure}

We design an aspect-word interactive graph (AWIG) with composed dependency tags to establish the relations between the given aspect term and potential long-distance opinion words for modeling global information. The construction process of an AWIG is shown in Figure \ref{fig:comp}. Following the setting of \cite{DBLP:conf/acl/WangSYQW20}, we set the given aspect term as a central node, and unlike the local encoder, we merge the given aspect embedding into a unique node $\bm{h_a}$ by summing it up if one aspect term consists of multiple tokens. If the input sentence contains more than one aspect term, we construct a unique AWIG for each aspect term to reduce the impact of irrelevant words. And we directly connect the given aspect term and the context words that both one-hop and multi-hop connections with the aspect term in the original dependency parse tree produced by the Biaffine Parser \cite{DBLP:conf/iclr/DozatM17}. By this means, we move the foreign opinion words around aspect terms, which is beneficial to model global information. The hidden contextual representations from BERT are composed of word embeddings $\{\bm{h_1}, \bm{h_2}, ..., \bm{h_n}\}$ and the given aspect embedding $\{\bm{h_i^a}, \bm{h_{i+1}^a}, ..., \bm{h_{i+\tau}^a}\}$, and $n$ is the number of tokens in the input sentence. We select the embeddings of context words serving as the features of corresponding context word nodes, and the given aspect embedding plays the role of the aspect node feature. When we build AWIG, we discard the words not connected to the given aspect term in the original dependency parse tree regardless of one hop or multiple hops. Thus, the number of word nodes in AWIG, we denote it as $m$, is not equal to $n$.

\label{section: section 3.3.1}
Then, we tend to assign the dependency relation representations $\{\bm{e_{a1}}, \bm{e_{a2}}, ..., \bm{e_{am}}\}$ to the corresponding edges since AWIG is edge attributed. Unlike \cite{DBLP:conf/acl/WangSYQW20}, which discards the dependency relations that do not connect to the aspect term node directly, we design a composed dependency tag to represent the relation from the aspect term to each neighbor node so that the features of multi-hop dependency relations are no longer discarded. Concretely, the form of the composed dependency tag is $r_{am}$ = $[dep_1:dep_2:...:dep_\kappa:\kappa]$, where $\kappa$ represents the distance between aspect term node $a$ and neighbor node $m$ in the original dependency parse tree, and $dep_\kappa$ represents the original dependency tag between two nodes generated by the Biaffine Parser \cite{DBLP:conf/iclr/DozatM17}, which is shown in Figure \ref{fig:comp}. Then, we feed $r_{am}$ = $[dep_1:dep_2:...:dep_\kappa:\kappa]$ into dependency embedding layer to get $\bm{e_{am}}$, and $\bm{e_{am}}$ is the concatenation of all the embeddings of elements in $r_{am}$. After AWIG is constructed, we feed the AWIG attached with its node and edge features into DGAT.

\subsubsection{Dual-level graph attention network (DGAT)}
\begin{figure}
    \centering
    \includegraphics[width=0.5\textwidth]{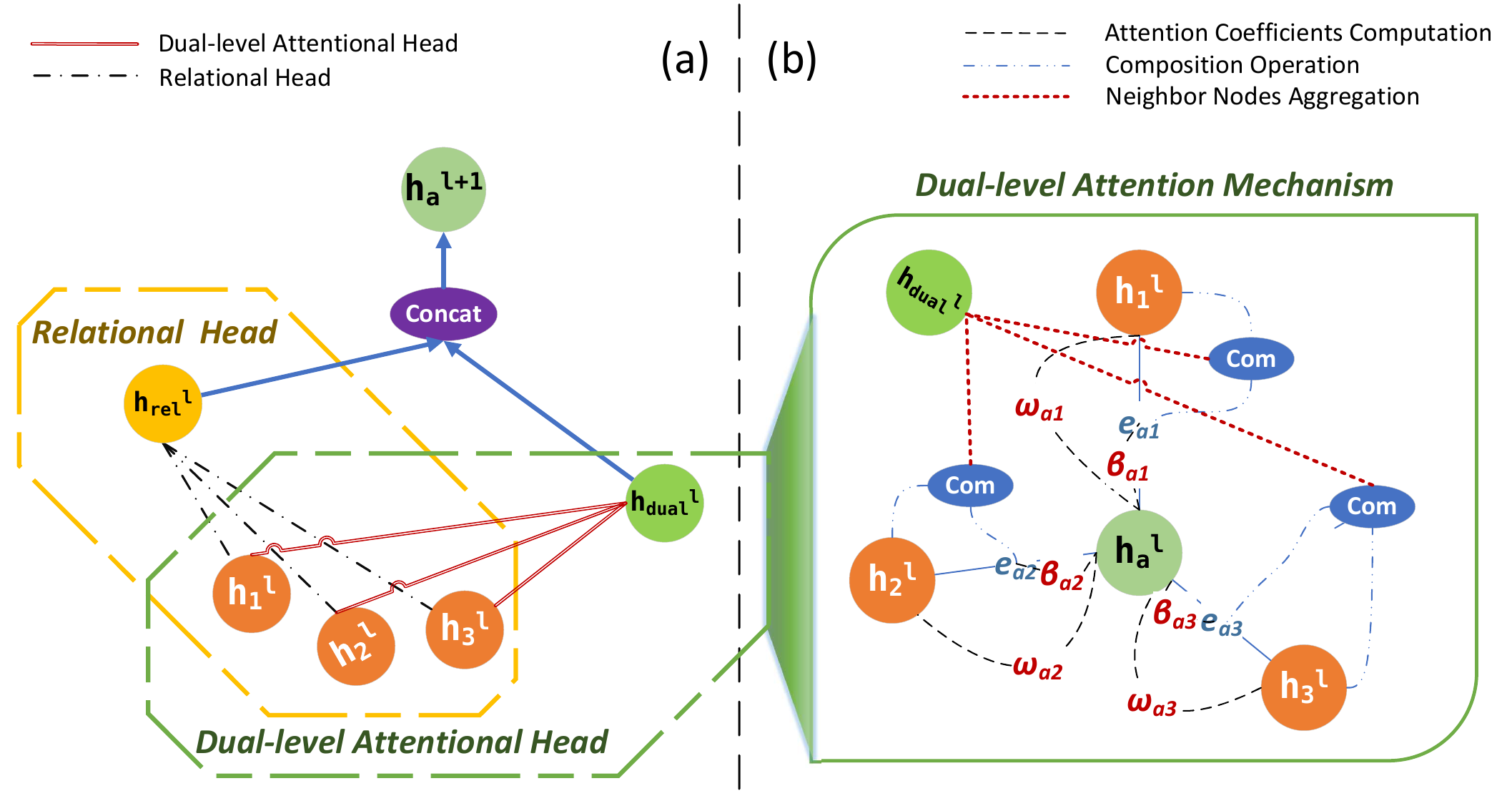}
    \caption{Overview of the proposed DGAT (a). DGAT comprises two sorts of multi-heads attention mechanism: dual-level attention head (b) and relational head.}
    \label{fig:DGAT}
\end{figure}

Intuitively, neighbor nodes with different dependency edges connected to the target node could have a distinct contribution to the target node. Thus, the target node should focus on different dependency edges before paying attention to the connected nodes. We design the DGAT based on this observation. The overall architecture of DGAT is shown in Figure \ref{fig:DGAT}(a). 
DGAT is composed of two sorts of multi-heads attention mechanism: dual-level attention head and relational head. The dual-level attention mechanism is demonstrated in Figure \ref{fig:DGAT}(b). We illustrate the dual-level attention mechanism by split it into three folds.

\paragraph{Attention Coefficients Computation:}
 Specifically, the novel part of our approach is to map dependency relation tags $r_{ai}$ into vector representations, denoted as $\bm{e_{ai}}$, which represents the relation embedding between aspect term $a$ and context word $i$. We get the representation of each word, denoted as $\bm{h_i}$. The $l$-th layer of the first level attention, target-edge attention, is computed as follow:
\begin{equation}
    \beta_{ai}^{lu} = softmax([\bm{W_a^lh_a^l}][\bm{W_e^le_{ai}^l}]),
\end{equation}
where $\bm{h_a^l}$ represents the embedding of aspect term $a$ at layer $l$, $\bm{e_{ai}^l}$ represents the dependency relation representation between aspect term $a$ and context word $i$ at layer $l$, and $\bm{W_a^l}$ and $\bm{W_e^l}$ are input transformation matrices for aspect term and relation embeddings at layer $l$, respectively. $\beta_{ai}^{lu}$ is a normalized target-edge attention coefficient computed by the u-th attention head at layer $l$.

The $l$-th layer of second level attention, target-node attention, is computed as follow:
\begin{equation}
    \omega_{ai}^{lu} = softmax(\beta_{ai}^{lu}[\bm{W_a^lh_a^l}][\bm{W_i^lh_i^l}]),
\end{equation}
where $\bm{h_i^l}$ represents the word embedding of context word $i$ at layer $l$, and $\bm{W_i^l}$ is a transformation matrix for context words at layer $l$.

\paragraph{Composition Operation:}
We also import cirular-correlation\cite{DBLP:conf/iclr/VashishthSNT20, DBLP:conf/aaai/NickelRP16} as a composition operation to incorporate the edge feature along with the node feature. The operation is computed as follow:
\begin{equation}
    \psi(\bm{h_i^l}, \bm{e_{ai}^l}) = \mathcal{F}^{-1}\left(\overline{\mathcal{F}(\bm{h_i^l})} \odot (\mathcal{F}(\bm{e_{ai}^l})\right),
\end{equation}
where $\psi$ is the cirular-correlation operation, and $\mathcal{F}(\cdot)$ and $\mathcal{F}^{-1}(\cdot)$ denote the Fast Fourier Transform (FFT) and its inverse. The $\overline{\mathcal{F}(\bm{h_i^l})}$ is the complex conjugate of $\mathcal{F}(\bm{h_i^l})$, and $\odot$ denotes the Hadamard product.

\paragraph{Neighbor Nodes Aggregation:}
Given a specific aspect term $a$, and the representation of its neighboring context word $\bm{h_i} \in \mathcal{N}_{a}$, the dual-level attention head is computed as:
\begin{equation}
    \bm{h_{dual}^{l+1}} = ||^U_{u=1}\sum_{i\in{\mathcal{N}_a}} \omega_{ai}^{lu} \psi(\bm{h_i^{l}}, \bm{e_{ai}^l}).
\end{equation}
Let $\bm{e_{ai}^{l+1}}$ denote the representation of $r_{ai}$ after $l$ layers. Then, 
\begin{equation}
    \bm{e_{ai}^{l+1}} = \bm{W_{r}^{l} e_{ai}^{l}}, 
\end{equation}
where $\bm{W_{r}^{l}}$ is a learnable transformation matrix to project dependency relation representations to the same embedding space as nodes.

Formally, DGAT contains U dual-level attention heads and V relational heads. The final representation of the aspect term $a$ is computed as follow:
\begin{equation}
    \bm{h_a^{l+1}} = \bm{h_{dual}^{l+1}} \;||\; \bm{h_{rel}^{l+1}},
\end{equation}
where $\bm{h_{rel}^{l+1}}$ is the relational head \cite{DBLP:conf/acl/WangSYQW20} computed as follows.
\begin{gather}
    \bm{h_{rel}^{l+1}} = ||^V_{v=1}\sum_{i\in{\mathcal{N}_a}}\rho_{ai}^{lv}\bm{W_v^l}\bm{h_i^l}, \\
    \rho_{ai}^{lv} = softmax(relu(\bm{e_{ai}^l}\bm{W_{v1}} + \bm{b_{v1}})\bm{W_{v2}} + \bm{b_{v2}}),
\end{gather}
where $\bm{W_v^l}$, $\bm{W_{v1}}$, $\bm{W_{v2}}$, $\bm{b_{v1}}$ and $\bm{b_{v2}}$ are learnable weights and biases, respectively. 

\subsection{Output Layer}
In the end, we concatenate both the local and global output features as the final representation $\bm{h_{final}}$. Then, we feed $\bm{h_{final}}$ into a fully connected layer and map it to probabilities over the different sentiment polarities.
\begin{equation}
    P(y=c) = softmax(\bm{W_Ph_{final}} + \bm{b_P}),
\end{equation}
where $\bm{W_P}$ and $\bm{b_P}$ are the weight matrix and bias, respectively. $P \in \mathcal{R}^C$ is the probability distribution for the sentiment polarity of a specific aspect term, where $C$ is the set of sentiment classes. The training objective is to minimize the standard cross-entropy loss with L2-regularization:
\begin{equation}
    L(\Theta) = - \sum_{(S,A) \in D}log(y=c) + \Lambda||\Theta||_2,
\end{equation}
where $D$ is the set of training data, $\Theta$ represents all trainable parameters, and $\Lambda$ is the coefficient of the L2-regularization term.

\section{Experiments}
\subsection{Datasets and Experimental Setup}
To verify the effectiveness of our proposed model, we conduct experiments on SemEval 2014 task 4 \cite{pontiki-etal-2014-semeval} and Twitter \cite{DBLP:conf/acl/DongWTTZX14} datasets. The SemEval 2014 dataset consists of Laptop and Restaurant datasets, and we remove the conflict category to alleviate the imbalance of these two datasets. And all of the three datasets contain three label category: \textsc{positive}, \textsc{negative} and \textsc{neutral}. The details of the experimental datasets are shown in Table \ref{tab:dataset}.

\begin{table}
\centering
\caption{Statistics of the SemEval 2014 and Twitter datasets.}
\label{tab:freq}
\scalebox{0.8}{
\begin{tabular}{c|cc cc cc}
\hline
\multirow{2}{*}{Dataset} & \multicolumn{2}{c}{Positive} & \multicolumn{2}{c}{Neutral} & \multicolumn{2}{c}{Negative} \\
& Train & Test & Train & Test & Train & Test \\
\hline
\hline
Restaurant & 2164 & 728 & 807 & 196 & 637 & 196 \\
Laptop & 994 & 341 & 870 & 128 & 464 & 169 \\
Twitter & 1561 & 173 & 3127 & 346 & 1560 & 173 \\
\hline
\end{tabular}}
\label{tab:dataset}
\end{table}

We utilize the last hidden states of the pre-trained BERT-base model for word representations \cite{DBLP:conf/naacl/DevlinCLT19}, the BERT containing 12 hidden layers, and 768 hidden dimensions for each layer. And the dimension of the dependency relation tag embedding is 768. The hidden dimension of DGAT is 128, and the number of heads is 6. The sample interval of Gaussian mask layer is 0.2. The dropout rate is 0.8. The number of the epoch is 30. We use Adam optimizer \cite{DBLP:journals/corr/KingmaB14} while training with the learning rate initialized by 0.00005. Our code will be released later.

\subsection{Baselines}
To testify the effectiveness of GDD, we select representative methods from the following two technical routes, which are listed below:


\paragraph{\textbf{Attention-based models:}}
\textbf{ATAE-LSTM} \cite{DBLP:conf/emnlp/WangHZZ16} tends to combine learned attention embeddings with LSTM to make predictions.

\textbf{IAN} \cite{DBLP:conf/ijcai/MaLZW17} is introduced to learn the coarse-grained attentions between the contexts and aspects interactively.

\textbf{RAM} \cite{DBLP:conf/emnlp/ChenSBY17} proposes to learn multi-hop attention based on BiLSTM and utilize GRU to obtain aggregated vector for prediction.
    
\textbf{MGAN} \cite{DBLP:conf/emnlp/FanFZ18} uses a two-level attention, fine-grained and coarse-grained, to capture the interaction between aspects and contexts.

\textbf{BERT} \cite{DBLP:conf/naacl/DevlinCLT19} fine-tunes BERT model to predict the sentiment polarity of the given aspect term.
    
\textbf{AEN-BERT} \cite{DBLP:journals/corr/abs-1902-09314} adopts attention mechanism and BERT to model the relationships between context words and aspect terms. 
    
\textbf{RPAEN-BERT} \cite{DBLP:journals/kais/WuXGLYW21} fine-tunes BERT and incorporates relative position information and aspect attention into model for ASC.
    
\textbf{LCF-BERT-CDW} \cite{zeng2019lcf} applys a context feature weight/mask to alleviate the influence of irrelevant words.
    
\textbf{LCFS-ASC-CDW} \cite{DBLP:conf/acl/PhanO20} adjusts the syntactic relative distance by setting hyper-parameters to deemphasize the adverse effects of unrelated words. 

\textbf{BERT-ADA} \cite{DBLP:conf/lrec/RietzlerSOE20} proposes a domain-adapted BERT for the Restaurant and Laptop datasets and obtains a promising performance.

\paragraph{\textbf{Graph-based models:}}
\textbf{GAT} \cite{huang-carley-2019-syntax} tends to utilize GAT to capture syntactic dependency relationships between words and aspects.

\textbf{ASGCN} \cite{zhang-etal-2019-aspect} combines BiLSTM to capture contextual information regarding word orders with a multi-layered GCN for prediction.

\textbf{CDT} \cite{sun-etal-2019-aspect} learns representations of sentences using BiLSTM, and further enhance the representations with GCN on dependency 
tree.

\textbf{TD-GAT} \cite{huang-carley-2019-syntax} incorporates a LSTM unit in the GAT and employs it on the syntactic dependency graph for prediction.

\textbf{BERT-RGAT} \cite{DBLP:conf/acl/WangSYQW20} feeds a reshaped dependency tree-based graph into RGAT to capture long-distance dependency information.

\begin{table}
  \centering
  \label{tab:freq}
  \caption{ Overall performance of different methods on SemEval 2014 task 4 and Twitter. The best scores are in bold. The results indicated by an asterisk(*) are reproduced following the methodology in the published paper. The other results except for our model are from the results reported by other baseline papers.}
  \scalebox{0.7}{
   \begin{tabular}{c|cccc cccc cccc}
    \hline
    \multirow{2}{*}{Model}    &\multicolumn{2}{c}{Restaurant}
    &\multicolumn{2}{c}{Laptop} &\multicolumn{2}{c}{Twitter} \cr
    &$\textit{Accuracy}$ &$\textit{Macro-F1}$&$\textit{Accuracy}$ &$\textit{Macro-F1}$&$\textit{Accuracy}$ &$\textit{Macro-F1}$ \\
    \hline 
    \hline
    ATAE-LSTM  &77.20 &- &68.70 &- & - & - \\
    IAN  &78.60 &- &72.10 &- & - & - \\ 
    RAM &80.23 &70.80 &74.49 &71.35 & 69.36 & 67.30\\
    MGAN &81.25 &71.94 &75.39 &72.47 & 72.54 & 70.81 \\
    \hline
    \hline
    GAT &78.21 &67.17 &73.04 &68.11 & 71.67 & 70.13 \\
    ASGCN &80.77 &72.02 &75.55 &71.05 & 72.15 & 70.40 \\
    CDT &82.30 &74.02 &77.10 &72.99 & 74.66 & 73.66 \\
    TD-GAT &80.35 &76.13 &74.13 &72.01 & 72.68 & 71.15 \\
    \hline
    \hline
     BERT &85.62 &78.28 &77.58 &72.38 & 75.28 & 74.11\\
     AEN-BERT & 83.12 & 73.76 & 79.73 & 76.31 & 74.71 & 73.13\\
     RPAEN-BERT & 85.20 & 78.00 & 80.60 & 76.80 & 74.80 & 72.90\\
     LCF-BERT-CDW* & 85.91 & 79.12 & 80.21 & 76.20 & - & -\\
     LCFS-ASC-CDW &86.71 &80.31 &80.52 &77.13 & - & - \\
     BERT-ADA & 87.14 & 80.05 & 79.19 & 74.18 & - & - \\
     BERT-RGAT & 86.60 & 81.35 & 78.12 & 74.07 & 76.15 & 74.88 \\
    GDD &\textbf{87.50} &\textbf{82.31} &\textbf{81.74} &\textbf{78.21} &\textbf{76.34} &\textbf{75.09}\\
    \hline
  \end{tabular}}
  \label{tab:overall}
 \end{table}

\subsection{Overall Performance}
The overall performance of our method and all the compared methods are listed in Table \ref{tab:overall}. The main evaluation matrices are Accuracy and Macro-averaged F1-score. The results of LCF-BERT-CDW are reproduced following the methodology in the published paper, and the results of BERT-ADA are under the in-domain setting, which is the same as our setting. Generally speaking, our GDD outperforms all the baseline models related, not only attention-based but graph-based. Specifically, compared with LCFS-ASC-CDW and BERT-RGAT, GDD can simultaneously focus on local information with the help of Gaussian mask as well as covariance self-attention and capture global information that the syntactic dependency tree conveyed by DGAT. Hence, our GDD outperforms these two compared methods. On the other hand, RPAEN-BERT captures local information around aspect terms by applying a relative position encoding, but RPAEN-BERT is also inferior to our GDD. Thus, the method proposed is effective in enhancing performance.

\begin{table}
\centering
\label{tab:freq}
\caption{Results of the ablation study.}
\scalebox{0.6}{
\begin{tabular}{c|cccc cccc cccc}
\hline
    \multirow{2}{*}{Model}    &\multicolumn{2}{c}{Restaurant}   &\multicolumn{2}{c}{Laptop}&\multicolumn{2}{c}{Twitter}\cr
    &$\textit{Accuracy}$ &$\textit{Macro-F1}$&$\textit{Accuracy}$ &$\textit{Macro-F1}$&$\textit{Accuracy}$ &$\textit{Macro-F1}$ \\
    \hline
    \hline
    GDD&{\textbf{87.50}}&\textbf{82.31}&\textbf{81.74}&\textbf{78.21}&\textbf{76.34}&\textbf{75.09}\\
    
    w/o Gaussian Mask & 85.98 & 79.41 & 80.10 & 76.45 & 74.11 & 72.46 \\
    w/o Covariance SA(+Original SA) &85.54 & 78.31 & 80.26 & 76.19 & 74.55 & 73.59\\
    w/o DGAT(+RGAT) & 85.18 & 78.83 & 80.43 & 77.47 & 74.55 & 72.63 \\
\hline
\end{tabular}}
\label{tab:ablation}
\end{table}

\subsection{Ablation Study}
We further conduct an ablation study to examine the effects of the different modules we designed. In Table \ref{tab:ablation}, we investigate our typical ablation conditions. The method \emph{w/o Gaussian Mask} means that we remove the Gaussian mask layer, and only the covariance self-attention layer is left in the local encoder. We can find out that without capture local information by Gaussian mask layer, the performance is declined on these three datasets. The method \emph{w/o Covariance SA(+Original SA)} denotes that we replace the covariance self-attention layer with the original self-attention layer, and the result of \emph{w/o Covariance SA(+Original SA)} is inferior to the overall GDD method. We can draw conclusions that the covariance self-attention layer is critical for our model. The method \emph{w/o DGAT(+RGAT)} means that we remove the DGAT and only use vanilla RGAT to encode global information, and the performance is also degraded. The result shows that our DGAT can utilize edge information better and enhance the performance of GDD.

\begin{figure}
    \centering
    \includegraphics[width=0.5\textwidth]{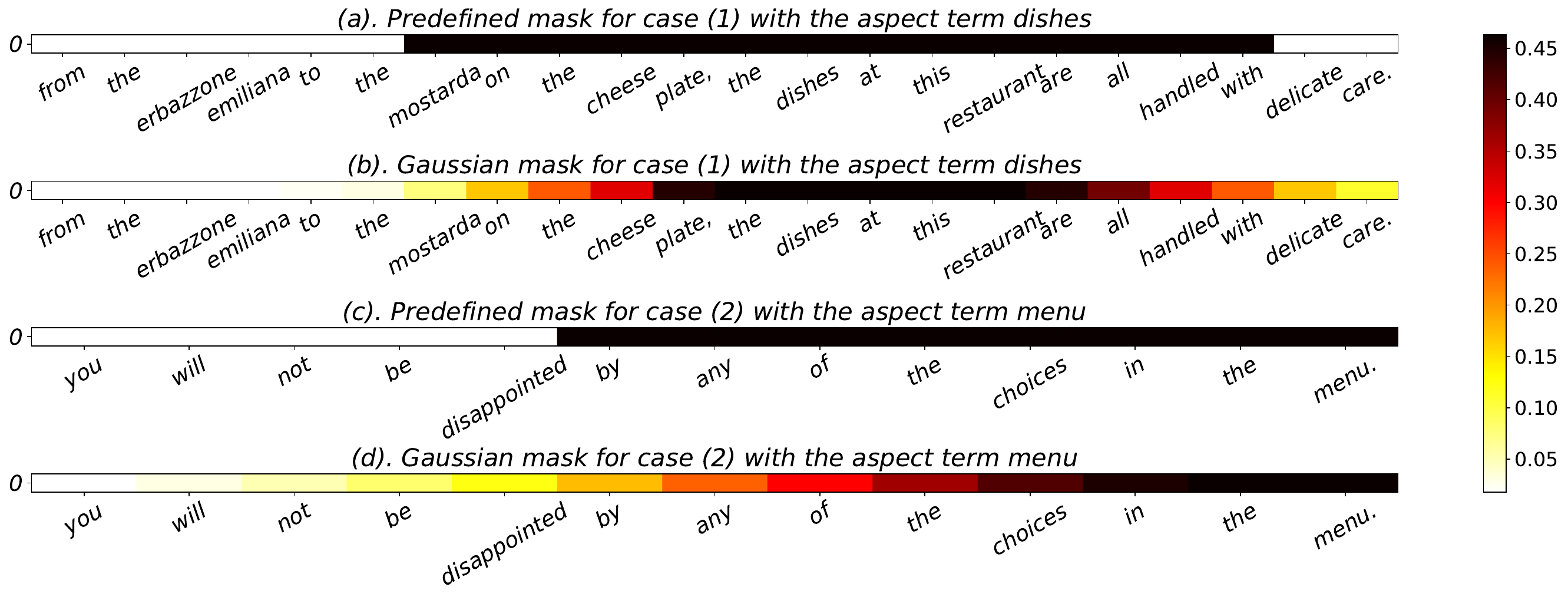}
    \caption{Gaussian mask visualization of typical cases.}
    \label{fig:Gaussian}
\end{figure}

\subsection{Visualization of Gaussian Mask}
To qualitatively demonstrate the effectiveness of our Gaussian mask layer, we conduct comparative experiments by using a predefined mask layer \cite{DBLP:conf/acl/PhanO20} instead of the Gaussian mask layer, and we set the value of the local context receptive field as 7 following the recommendation of \cite{zeng2019lcf}. We visualize the Gaussian masks and the predefined masks by demonstrating two typical cases in Figure \ref{fig:Gaussian}, and the white area indicates that the mask value is zero. Figure \ref{fig:Gaussian}(a) and (b) display the case (1): \textit{From the erbazzone emiliana to the mostarda on the cheese plate, the dishes at this restaurant are all handled with delicate care.} with the given aspect term \textit{dishes}. We can find out that the Gaussian mask can adaptively adjust the receptive field surround the given aspect term to capture critical information that the words \textit{delicate care} conveyed. Whereas the predefined mask misses the critical information since it down-weights the contribution of words to zero that are out of the local context receptive field, which leads to prediction failure. Indeed, we could adjust hyper-parameters to achieve the same purpose, but these hyper-parameters can not adapt to all situations. Figure \ref{fig:Gaussian}(c) and (d) display the case (2) \textit{You will not be disappointed by any of the choices in the menu.} with the given aspect term \textit{menu}. We can discover that the local context receptive field setting by hyper-parameters is not suitable for this case either, but our Gaussian mask adjusts its receptive field to cover the critical information \textit{not be disappointed} for predicting correctly. In addition, comparing Figure \ref{fig:Gaussian}(b) and (d), we can find that the scopes of receptive fields surround aspect terms \textit{dishes} and \textit{menu} are different, which indicates that the Gaussian mask can adjust its receptive field according to the different cases. These verify that our Gaussian mask layer can capture local features more adaptively and is more effective than the predefined mask \cite{DBLP:conf/acl/PhanO20,zeng2019lcf}. Thus, the local encoder armed with the Gaussian mask layer can mine local information flexibly and effectively.

\begin{figure}
\begin{minipage}[t]{0.4\textwidth}
    \centering
    \includegraphics[width=1.0\textwidth]{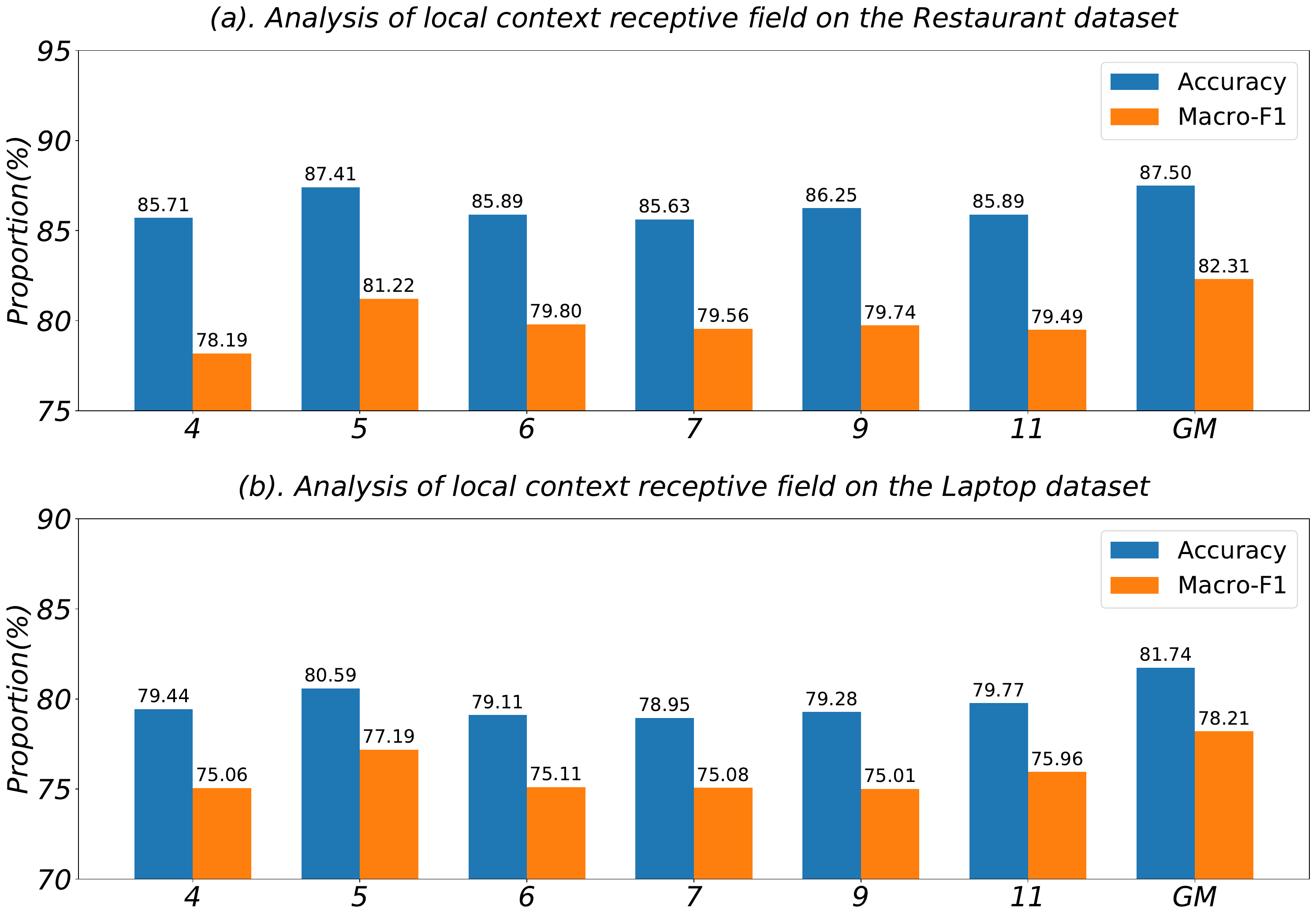}
    \caption{Analysis of local context receptive field on the Restaurant and Laptop datasets.}
    \label{fig:receptive}
\end{minipage}
\end{figure}

\subsection{Analysis of Local Context Receptive Field}
To further verify the effectiveness of our Gaussian mask layer, we analyze the impacts of different local context receptive fields on the performance of GDD on the Restaurant and Laptop datasets. The results are illustrated in Figure \ref{fig:receptive}, and the evaluation matrices are Accuracy and Macro-averaged F1-score. We employ the predefined mask proposed by \cite{DBLP:conf/acl/PhanO20} instead of our Gaussian mask on our model, and the values of local context receptive field are \textit{4, 5, 6, 7, 9, 11} severally for the predefined mask. The annotations on the x-axis represent the local context receptive field values, and 'GM' denotes the result by employing the Gaussian mask layer. We can observe that when the value of the receptive field is 5, this performance is the best among the results of the predefined mask on these two datasets, but these results are still inferior to the performances of employing the Gaussian mask layer on GDD. These results demonstrate that our Gaussian mask layer is more effective than the predefined mask and beneficial to improve the performance of GDD.

\subsection{Attention Distribution Exploration of Covariance Self-attention}
\begin{figure}[htbp]
\centering
\subfigure[The attention matrix of the input case generated by the covariance self-attention layer.]{
\begin{minipage}[t]{0.22\textwidth}
\centering
\includegraphics[width=0.9\textwidth]{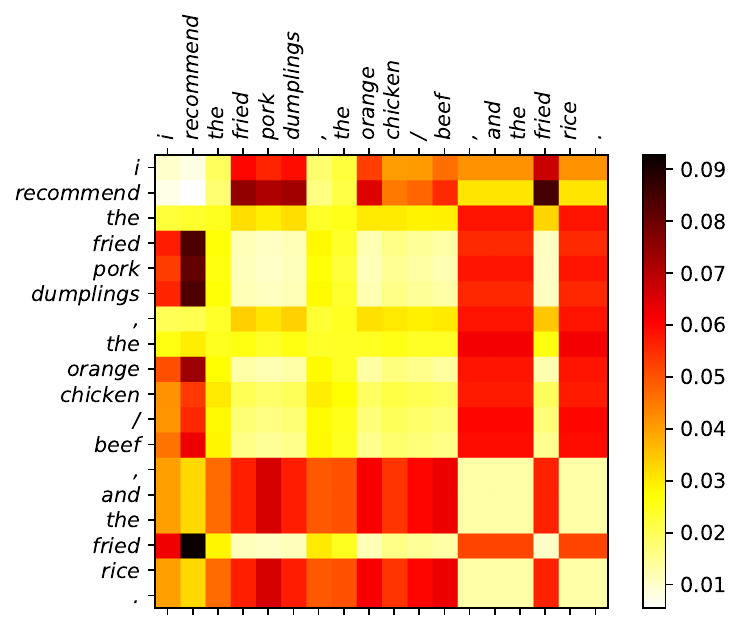}
\end{minipage}
}
\subfigure[The attention matrix of the input case generated by the original self-attention layer.]{
\begin{minipage}[t]{0.22\textwidth}
\centering
\includegraphics[width=0.9\textwidth]{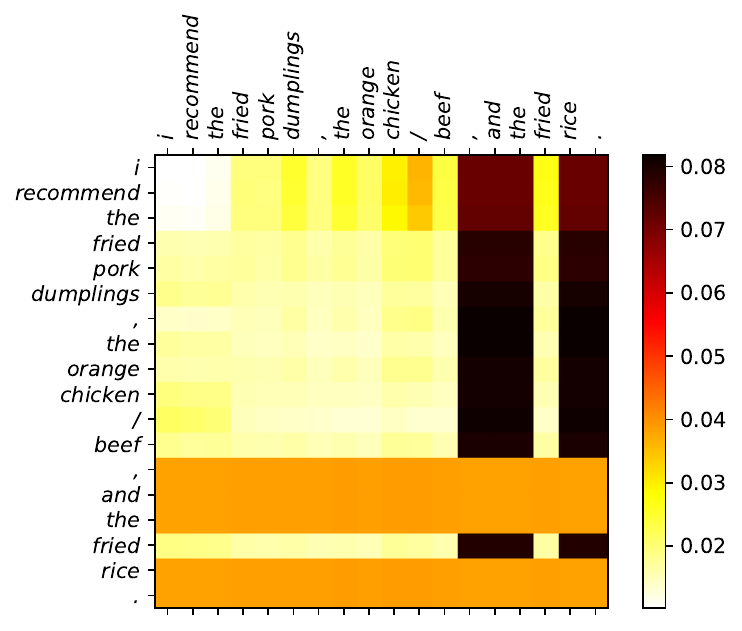}
\end{minipage}
}
\caption{The attention visualization of covariance self-attention and original self-attention.}
\label{fig:attention}
\end{figure}

To further figure out the effect of our covariance self-attention, we demonstrate the attention distribution visualization of covariance self-attention (a) and original self-attention (b) in Figure \ref{fig:attention}. The input sentence is \textit{I recommend the fried pork dumplings, the orange chicken/beef, and the fried rice}. And the given aspect term is \textit{fried rice} with the sentiment polarity label \textsc{Positive}. In this case, GDD with the original self-attention cannot obtain distinguishable attention scores notably, which leads to an incorrect prediction \textsc{Neural}. In contrast, GDD with the covariance self-attention can distinguish the opinion words more obviously and pay more attention to valuable words like \textit{recommend}, which leads to making a correct prediction \textsc{Positive}. Thus, the covariance self-attention can make attention weights more distinguishable than the original self-attention, which is helpful to make correct predictions.

\subsection{Attention Distribution Exploration of DGAT}
\begin{figure}
    \centering
    \includegraphics[width=0.5\textwidth]{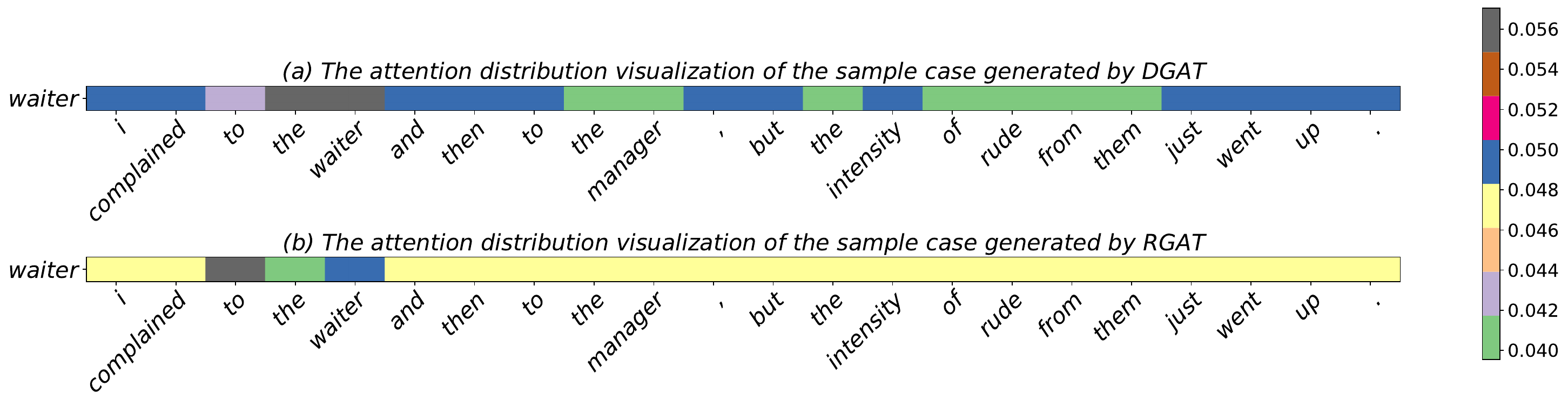}
    \caption{The attention distribution visualization of DGAT and RGAT.}
    \label{fig:DGAT_attention}
\end{figure}

In this section, to further testify the effectiveness of our DGAT, we illustrate the attention visualizations of DGAT and its compared method RGAT, which is displayed in Figure \ref{fig:DGAT_attention}. The sample case is \textit{I complained to the waiter and then to the manager, but the intensity of rude from them just went up}. And the given aspect term is \textit{waiter} attached with ground truth sentiment polarity \textsc{negative}. GDD equipped with DGAT makes the correct prediction, whereas GDD equipped with RGAT gets it wrong. Figure \ref{fig:DGAT_attention}(a) displays the attention distribution generated by DGAT, we can observe that our DGAT can recognize valuable tokens such as \textit{complained}, \textit{but} and \textit{went up}, and the attention weights are more discriminative than RGAT. However, GDD equipped with RGAT can not clearly pick out valuable tokens, and the attention weights are almost the same, leading to prediction failure. Thus, our DGAT is essential to make the right prediction.

\subsection{Case Study}
We accumulate test cases from the SemEval 2014 Restaurant dataset to confirm the performance of GDD. The two top-performing baselines in Table \ref{tab:case study}, BERT-RGAT \cite{DBLP:conf/acl/WangSYQW20} and LCFS-ASC-CDW \cite{DBLP:conf/acl/PhanO20}, are selected to compare with GDD. 
As shown in Table \ref{tab:case study}, GDD is correct on all 4 cases, while BERT-RGAT and LCFS-ASC-CDW are wrong on all these cases. In Table \ref{tab:case study}, cases (1)(2) do not contain prominent opinion words, whereas the whole indeed expresses a specific sentiment polarity on the given aspect. GDD can understand the semantics and make correct predictions. In cases (3)(4), despite existing opinion words, the sentences are so complex that the baseline models fail to capture the genuine sentiment behind texts. GDD can understand the sentence's implicit sentiment and predict correctly by incorporating local and global information. All above shows that GDD is better than these two compared models and can deal with complicated cases to a certain extend.

\begin{table}
    \centering
    \caption{Sample cases extracted from the Restaurant dataset, and the predicted results of GDD, BERT-RGAT, LCFS-ASC-CDW, as well as the corresponding ground truth.}
    \scalebox{0.53}{
    \begin{tabular}{c|c|c||c|c|c|c}
    \hline
    \multirow{2}{*}{} & \multicolumn{2}{c||}{\textbf{Sample Cases}} & \multicolumn{4}{c}{\textbf{Sentiment Polarity}} \\
    \cline{2-7}
    & $\textit{Sentence}$ & $\textit{Aspect}$ & $\textit{Ground truth}$ & $\textit{BERT-RGAT}$ & $\textit{LCFS-ASC-CDW}$ & $\textit{GDD}$ \\
    \hline
    \hline
    \multirow{2}{*}{(1).} & \textit{Go with the specials, and} & \multirow{2}{*}{\textit{specials}} & \multirow{2}{*}{\textsc{Positive}} & \multirow{2}{*}{\textsc{Neutral}} & \multirow{2}{*}{\textsc{Negative}} & \multirow{2}{*}{\textbf{\textsc{Positive}}$_\surd$} \\
    & \textit{stay away from the salmon.} &&&&& \\
    \hline
    \multirow{3}{*}{(2).} & \textit{For the price you pay for the food} & \multirow{3}{*}{\textit{food}} & \multirow{3}{*}{\textsc{negative}} & \multirow{3}{*}{\textsc{neutral}} & \multirow{3}{*}{\textsc{neutral}} & \multirow{3}{*}{\textbf{\textsc{negative}}$_\surd$} \\
    & \textit{here, you'd expect it to be at least} &&&&& \\
    & \textit{on par with other Japanese restaurants.} &&&&& \\
    \hline
    (3). & \textit{The staff should be a bit more friendly.} & \textit{staff} & \textsc{Negative} & \textsc{Positive} & \textsc{Neutral} & \textbf{\textsc{Negative}}$_\surd$ \\
    \hline
    \multirow{2}{*}{(4).} & \textit{Perfectly al dente pasta, not drowned} & \multirow{2}{*}{\textit{sauce}} & \multirow{2}{*}{\textsc{Neutral}} & \multirow{2}{*}{\textsc{Positive}} & \multirow{2}{*}{\textsc{Positive}} & \multirow{2}{*}{\textbf{\textsc{Neutral}}$_\surd$} \\
    & \textit{in sauce - - generous portions.} &&&&& \\
    \hline
    \end{tabular}}
    \label{tab:case study}
\end{table}

\section{Conclusions}

In this paper, we propose a novel and effective model GDD to tackle aspect-based sentiment classification. Thanks to the Gaussian mask layer and the covariance self-attention layer, GDD can acquire local context information adaptively. And with the help of DGAT, GDD excavates global information effectively. Our GDD can encode both local and global information simultaneously and promote the performance for ASC.
The results demonstrate that GDD outperforms existing models and achieves the state of the art performance.

\bibliographystyle{ACM-Reference-Format}
\bibliography{sample-sigconf}


\end{document}